# Indoor Drone Localization and Tracking Based on Acoustic Inertial Measurement

Yimiao Sun, *Student Member, IEEE*, Weiguo Wang, *Student Member, IEEE*, Luca Mottola, *Member, IEEE*, Jia Zhang, *Graduate Student Member, IEEE*, Ruijin Wang, *Member, IEEE*, and Yuan He, *Senior Member, IEEE*

*Abstract*—We present Acoustic Inertial Measurement (AIM), a one-of-a-kind technique for indoor drone localization and tracking. Indoor drone localization and tracking are arguably a crucial, yet unsolved challenge: in GPS-denied environments, existing approaches enjoy limited applicability, especially in Non-Line of Sight (NLoS), require extensive environment instrumentation, or demand considerable hardware/software changes on drones. In contrast, AIM exploits the acoustic characteristics of the drones to estimate their location and derive their motion, *even in NLoS* settings. We tame location estimation errors using a dedicated Kalman filter and the Interquartile Range rule (IQR) and demonstrate that AIM can support indoor spaces with arbitrary ranges and layouts. We implement AIM using an off-the-shelf microphone array and evaluate its performance with a commercial drone under varied settings. Results indicate that the mean localization error of AIM is 46% lower than that of commercial UWB-based systems in a complex 10 m×10 m indoor scenario, where state-of-the-art infrared systems would not even work because of NLoS situations. When distributed microphone arrays are deployed, the mean error can be reduced to less than 0.5m in a 20m range, and even support spaces with arbitrary ranges and layouts.

*Index Terms*—Acoustic signal, drone, indoor tracking, microphone array.

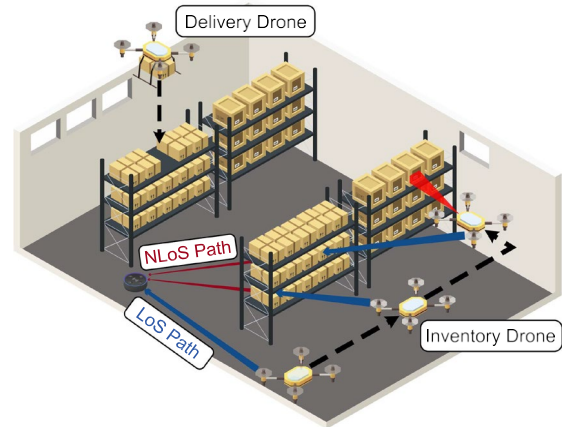

Fig. 1. Example of AIM's application scenario.

## I. INTRODUCTION

LOCATION information is crucial for drone operation [1], [2], regardless of the application and target deployment environment [3], [4], [5]. For example, in an indoor warehouse like the one of Fig. 1, a drone for cargo inventory needs location information to determine the position of the cargo relative to its own. When performing cargo deliveries, a drone must follow the predefined route and land at the right target location for the drop-off.

Location information must be *accurate*. Errors in location estimates may not just degrade system performance, but represent a safety hazard as the drone's own movements are largely determined by location information. In outdoor settings, GPS is arguably the mainstream to provide accurate location. The indoor setting, however, represents a completely different ballgame.

There have been many different approaches and solutions for drone localization and tracking [6], [7], [8], [9], [10]. Radar-based approaches [7], [11], for example, work both indoors and outdoors. Their spatial resolution is limited so that it is generally difficult to localize small-size drones. Further, objects in the target environment easily interfere with the radar signals, degrading the accuracy. RF-based localization approaches [9], [12] require installing wireless transceivers on the drone and reengineering the flight controller. Inertial measurement methods [13], [14] are useful when absolute localization is unavailable, but the accumulation of errors likely becomes an issue. Infrared-based systems require dedicated hardware and corresponding software changes on both drones and control stations [10].

A *low-cost* and *accurate* localization approach is arguably still missing on drones. Inspired by our observation on the dynamics of drones [15], [16], [17] and the existing work that utilizes the propellers to produce audio [18], we present Acoustic Inertial Measurement (AIM), a *completely passive* approach to localize the drones with a single microphone array. The term *passive*

Manuscript received 16 July 2023; revised 2 November 2023; accepted 15 November 2023. Date of publication 28 November 2023; date of current version 7 May 2024. This work was supported in part by the National Science Fund of China under Grants U21B2007 and 62271128, in part by the R&D Project of Key Core Technology and Generic Technology in Shanxi Province under Grant 2020XXX007, in part by the Swedish Science Foundation (SSF), the Digital Futures programme (project Drone Arena), the Swedish Research Council under Grant 2018-05024, and in part by KAW project UPDATE. Recommended for acceptance by C. M. Pinotti. *(Corresponding author: Yuan He.)*

Yimiao Sun, Weiguo Wang, Jia Zhang, and Yuan He are with the School of Software and BNRist, Tsinghua University, Beijing 100190, China (e-mail: sym21@mails.tsinghua.edu.cn; wwg18@mails.tsinghua.edu.cn; jzhang19@mails.tsinghua.edu.cn; heyuan@tsinghua.edu.cn).

Luca Mottola is with the Politecnico di Milano, 20133 Milano, Italy, also with the RI.SE , Sweden, and also with the Uppsala University, 751 05 Uppsala, Sweden (e-mail: luca.mottola@polimi.it).

Ruijin Wang is with the School of Computer Science and Engineering, University of Electronic Science and Technology of China, Chengdu 610056, China (e-mail: ruijinwang@uestc.edu.cn).

Digital Object Identifier 10.1109/TMC.2023.3335860

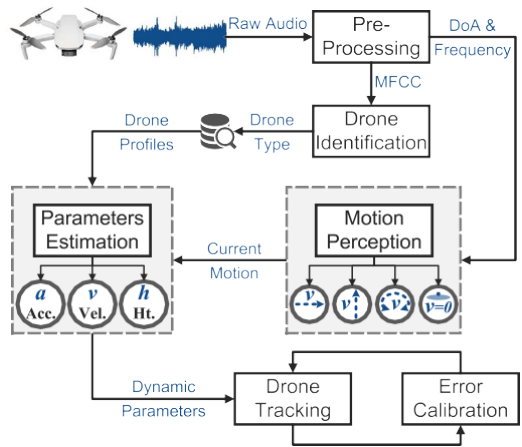

Fig. 2. *AIM* workflow.

means AIM requires *no* additional hardware and *no* software changes on the drones, only using the acoustic signals naturally produced by the drone itself. AIM works with only a single microphone array but may be extended with ease to support spaces with arbitrary ranges and layouts by deploying distributed arrays.

To achieve this, we must tackle three key challenges:
1) A single microphone array can only acquire one direction of arrival (DoA), which denotes the drone's direction relative to the array; this information alone is insufficient for location calculation.
2) The only input to AIM is the propellers' sound of the drone; how to infer the drone's location and motion from this single acoustic signal is an open problem.
3) In complex indoor environments, the acoustic channel between the drone and the microphone array is easily interfered by ambient noise and obstacles, or travels along NLoS paths as Fig. 1 illustrates.

*AIM:* We address these issues based on the fundamental observation that the rotating propellers create a *dual acoustic channel*: from the microphone array's view, the propellers are regarded as the sound source, so the DoA of sound denotes the *orientation* of the drone. At the same time, the propellers are also high-speed rotating machinery, so the frequency properties of the sound actually correspond to the rotating state of the propellers, which in turn determines the drone's *motion*. Obtaining *orientation and motion* information allows us to track the drone's location continuously.

Fig. 2 illustrates AIM's workflow. The raw acoustic signal captured by the microphone array is first pre-processed to extract the characteristics of the acoustic signal, for example, DoA, frequencies, and Mel-Frequency Cepstral Coefficients (MFCC). DoA and frequencies help deduce the drone's current motion, whereas MFCC is utilized for identifying the specific drone structure, for example, a quadcopter as opposed to an octocopter, and then loading the corresponding profile information (e.g., mass) from a database.

By feeding the drone's profiles into a set of dynamic equations we formulate, we estimate its dynamic parameters, that is, acceleration and velocity. The drone's location is calculated consequently. To reduce error, we adopt a dedicated Kalman filter and the Interquartile Range rule (IQR). We further show how AIM can be extended to support indoor spaces with arbitrary ranges and layouts by deploying distributed microphone arrays.

Our contribution can be summarized as follows:
1) We design AIM, a *completely passive* drone tracking approach that can work with a single microphone array. At the core of AIM is exploiting the dual acoustic channel to perceive the drone's motion and estimate its location.
2) We exploit the acoustic characteristics of the drones to derive their motion and estimate their location, *even in NLoS* settings. We combine this with a dedicated Kalman filter and the Interquartile Range rule (IQR) to reduce the error, and demonstrate that AIM can support indoor spaces with arbitrary ranges and layouts.
3) We implement AIM using off-the-shelf microphone arrays and perform an evaluation using a commercial drone under varied settings. Results indicate that the mean localization error of AIM in a complex 10 m $\times$ 10 m indoor scenario is 1.89 m, 46% lower than that of commercial UWB-based systems, where state-of-the-art infrared systems would not even work. Further, AIM can be extended to support indoor spaces with arbitrary ranges and layouts by deploying distributed microphone arrays.

Works close to our efforts are summarized in Section II. Section III introduces the unique acoustic features of different drone motions. Then, Section IV presents methods to distinguish different drone motions and drone structures. Section V elaborates on the core algorithm of AIM for drone trajectory tracking and Section VI unfolds how to use distributed microphone arrays to extend the operating range. The implementation and evaluation results are presented in Section VII. We discuss practical issues in Section VIII and conclude the paper in Section IX.

## II. RELATED WORK

The distinctive feature of our work is to perform drone localization and tracking using acoustic signals. We briefly survey existing efforts in either field.

### A. Drone Localization and Tracking

*RF-based methods:* RF signals are extensively explored for drone localization [9], [19], [20], [21], [22]. In outdoor scenarios, mmWave and WiFi are usually used. For example, mmHawkeye [20] exploits commercial mmWave radars to capture the feature of drone's periodic micro-motion (PMM) and achieve less than 10cm tracking error within 30m. Nguyen et al. [9] explore a passive approach to localize both the drone and its controller in 2.4 GHz WiFi frequency channel. They show an average error of around 10 m in the 30 m to 150 m distance.

In indoor scenarios, Ultra Wide Band (UWB)-based approaches are mainstream. UWB techniques [21], [23] achieve decimeter accuracy for drone tracking. To improve accuracy, UWB may integrate with other techniques, such as visual SLAM [24], RGB-D camera [25] and optical flows [26]. The errors of these methods are usually lower than 20m. However, the performance of RF-based methods will degrade in complex NLoS scenarios, especially in the presence of equipment that absorbs or scatters RF signal [27].

*Acoustics-based methods:* AIM enjoys the fact that acoustic signals may be fruitfully employed also in NLoS settings [28], [29]. For example, Mao et al. [30] attach two speakers on the drone to emit Frequency-Modulated Continuous-Wave (FMCW) signals, used to estimate the distance between the drone and a mobile phone. As for AIM, it does not install any extra equipment on the drone. Other efforts [31], [32] only regard the drone as a mobile sound source and deploy 3D or large microphone arrays to estimate its location. Compared with these techniques, we explore the theoretical connection between the drone's sound and its motions, deduce the drone's dynamic parameters, such as velocity and acceleration, from its sound and track the drone by using only a small 2D microphone array.

*Data-driven methods and other:* AIM is a model-driven technique for drone tracking and localization. Various data-driven methods exploiting machine learning or deep learning exist [33], [34], [35], [36]. However, these methods may require complex algorithms and pose challenges in transferring a specific model to another drone or environment, which makes them arguably impractical.

GPS is a mature approach widely used for drone localization and offers meter-level accuracy, but its application indoors is extremely difficult [37]. Methods based on optics and vision [10], [38], [39], [40] can provide much more accurate results for indoor drone localization, whose errors are even less than 1mm as reported [38]. However, these methods vastly assume line-of-sight (LoS) conditions and are sensitive to lighting conditions.

### B. Acoustics-Based Tracking

*Indoor tracking:* Several works demonstrate the use of acoustic signals for localization and tracking [41], [42]. With a single microphone array, Voloc [43] aligns the multi-path DoA estimation for accurate localization of indoor acoustic sources; Symphony [44] extends this method to localize multiple sources by leveraging the prior-known layout of the array. PACE [45] localizes multiple mobile users simultaneously by leveraging structure-borne and air-borne footstep impact sounds. These works assume that the localization target and the microphone array are on the same plane or that the target's altitude is known, to solve a bi-dimensional localization problem. Differently, we exploit the signal feature in both the spatial and frequency domains, achieving *three-dimensional* localization with a single array.

*Short-range tracking:* Recent works adopt wearable devices for tracking, such as smartwatches and earphones. SoM [46] tracks the wrist using a smartwatch with IMUs and employs the smartphone to send beacons for error calibration. Ear-AR [47] uses the IMU in earphones and smartphones to track the indoor user's location and gazing orientation. When the embedded microphone and speaker in the wired or wireless earphones have already formed a transceiver pair, EarphoneTrack [48] proposes to track either the microphone or speaker with this pair. Unlike what we do with AIM, these approaches are effective only in the short range, specifically between wearable devices and users' smartphones.

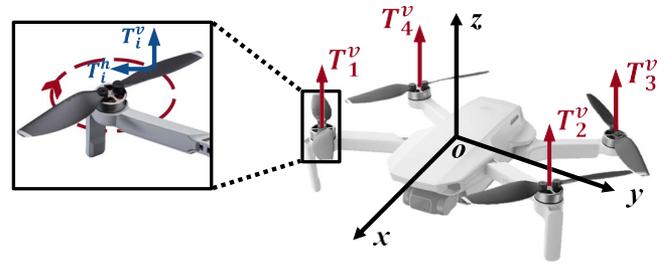

Fig. 3. Quadcopter drone structure.

## III. THE SOUND OF DRONES

In this section, we explore the features of a drone's sound signals and how they relate to motion.

### A. Key Features

Drone propellers are designed to displace the air around them. The resulting pressure gradient creates a force vector. We model the connection between the sound of the drone's propellers and its physical structure.

Fig. 3 illustrates the most common drone structure, that is, a quadcopter composed of two orthogonal arms. A propeller is mounted at either end of each arm. The force vector obtained by the propeller rotation can be decomposed into a vertical component $T_i^v$ and a horizontal component $T_i^h$.

The vertical component lifts the drone and can be calculated as $T_i^v = k^v f_i^2$, where $f_i$ is the rotation frequency of the $i$th propeller and $k^v$ is a constant related to the lift coefficient. The drag force $T_i^h$ horizontally controls the rotation of the body and can be calculated as $T_i^h = k^h f_i^2$, where $k^h$ is a constant related to drag coefficient [49]. The lift forces of all propellers follow the same direction, while the drag forces of adjacent propellers are opposite to compensate for the torque otherwise generated, which induces spinning.

The sound produced by the propellers is highly correlated with the frequency $f_i$ of each motor. Because each propeller has multiple blades, two in most cases, the fundamental frequency of the sound is not the rotation frequency $f_i$, but the blade passing frequency (BPF). The BPF is defined as $f_i^{BPF} = n f_i$, where $n$ is the number of blades. In addition to the BPF, harmonic frequencies may also be observed as an integer multiple of the BPF [50].

If we can capture the drone's sound and obtain the BPF as well as its harmonics, we may then estimate the rotation frequencies $f_i$, and thus the forces exerted by each propeller. Using a model of the drone's physical dynamics, which is necessarily a function of its mechanical structure, we may also estimate its direction and motion. This is the essence of the frequency-based localization and tracking in AIM.

### B. Sound and Motion

We analyze here the inner relationship between the drone's sound and its physical motion.

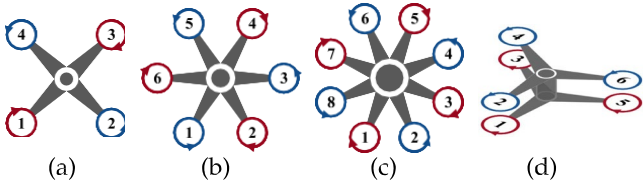

Fig. 4. Typical structures of four drone types: (a) quadcopter; (b) hexacopter; (c) octocopter; and (d) Y6. Different colors represent different directions of rotation.

TABLE I
CLASSIFICATION SCHEME OF THE FOUR MOTIONS

|  | Single-Peak | Multiple-Peak |
|---|---|---|
| Unstable DoA | Vertical linear motion | Horizontal linear motion |
| Stable DoA | Hovering motion | Yaw motion |

We theoretically analyze the acoustic properties of four common drone structures, shown in Fig. 4. Drone flights are composed of four basic motions: hovering, yaw, horizontal linear motion and vertical linear motion, as depicted in Fig. 5. Interestingly, we find that these basic motions exhibit different acoustic properties in the frequency domain because they are performed by changing each motor's rotation frequency $f_i$ differently. In the following, $N = 4, 6$ or $8$ depending on the drone structure among the ones in Fig. 4.

*Hovering:* in the absence of environmental effects requiring compensation, all propellers rotate at the same frequency to maintain the vertical and horizontal balance, so the drone remains stationary. Therefore, we have $f_i = f_j, 1 \leq i,j \leq N$.

*Yaw:* propellers operate in pairs, shown by different colors in Fig. 4. Each pair rotates at the same frequency, creating a rotational momentum while maintaining the vertical balance, which makes the drone rotate around the center. Thus, we have $f_{2i-1} = f_{2j-1} \neq f_{2i} = f_{2j}, 1 \leq i,j \leq \frac{N}{2}$.

*Horizontal motion:* propellers operate in pairs again, this time to tilt the body while maintaining the vertical balance. Then the drone moves horizontally. We use parentheses to indicate equal frequencies for brevity. When the drone tilts forwards or backwards, that is, it pitches, we have $(f_1 f_2) (f_3 f_4)$ for quadcopters, $(f_1 f_2) (f_3 f_6) (f_4 f_5)$ for hexacopters, $(f_1 f_2) (f_3 f_8) (f_4 f_7) (f_5 f_6)$ for octocopters and $(f_3 f_4 f_5 f_6) (f_1 f_2)$ for Y6 structures. Symmetric observations apply when the drone tilts leftwards or rightwards, that is, it rolls.

*Vertical motion:* all propellers rotate at the same speed to generate thrust greater or lower than the force of gravity on the drone. Accordingly, the drone moves upwards or downwards, so we have $f_i = f_j, 1 \leq i,j \leq N$.

In the following, we illustrate how these observations may be a stepping stone to achieving accurate drone localization and tracking.

## IV. MOTION AND STRUCTURE

We use the features of the sound signal in the frequency, spatial, and time domains to estimate the drone's motion and identify its structure. These two components are the basis of our system.

### A. Motion Detection

Based on the analysis of Section III, we conduct a proof-of-concept experiment to check whether the four basic motions can be distinguished by the sound characteristics. In this experiment, we use a DJI Mini 2 quadcopter and a microphone to receive the acoustic signal.

Fig. 6 shows the spectrum of the acoustic signal corresponding to the motions of Fig. 5 and conforms to our understanding of the drone's dynamics. Specifically, we observe two peak fundamental frequencies in the case of yaw and horizontal motion. In comparison, there is only one peak fundamental frequency in the case of hovering and vertical motions.

Exclusively based on frequency domains, we can only classify the four motions into two categories, depending on the number of peak fundamental frequencies. To resolve this ambiguity, we leverage the spatial information of the sound. Crucially, we note that the drone spatial coordinates are stable during hovering or yaw, while they change during vertical or horizontal motion. The change in position may be detected by the sound's DoA, as elaborated in Section V-A. By combining the information obtained from the number of peak fundamental frequencies and DoA as shown in Table I, AIM can correctly discern the four basic motions.

Detecting the four basic motions is vastly sufficient to localize and track drones in a multitude of indoor drone applications, including most of those we mention in the Introduction. In indoor settings, for example, warehouses or smart factories, planning of robot movements—not just drones—is most often achieved by sequentially combining the four basic motions. This is beneficial in at least two respects: i) it matches the regular physical layout of the target deployment scenarios; in a warehouse, for example, shelves are side-by-side horizontally laid and goods are stacked vertically; and ii) it greatly simplifies path planning, yielding much more scalable systems.

To further improve the accuracy in detecting the four basic drone motions, we further observe that high-frequency harmonics share similar characteristics with the fundamental frequencies. Because the noise in the low-frequency band is usually stronger than that in the high-frequency band, the harmonics may experience less noise than the original BPF. Thus, we estimate the BPF from the weighted average of both the fundamental frequencies and the harmonics, which are weighted by their amplitudes. For hovering, a single band is present on the spectrogram.

### B. Drone Structure Identification

There exist several types of drones apt to support distinct applications. For instance, drones with high load-carrying capacity can be designated to transport goods, while drones with large-capacity of batteries can be employed for environmental surveillance. Each such type of drone uses a different physical structure, expressly designed to optimize the aerodynamics features required to carry out a specific task. For AIM to work accurately, it is crucial to precisely recognize the particular drone structure once it is detected by the microphone array. In the following, we illustrate a technique to do so, even in case different drone types co-exist in the same area.

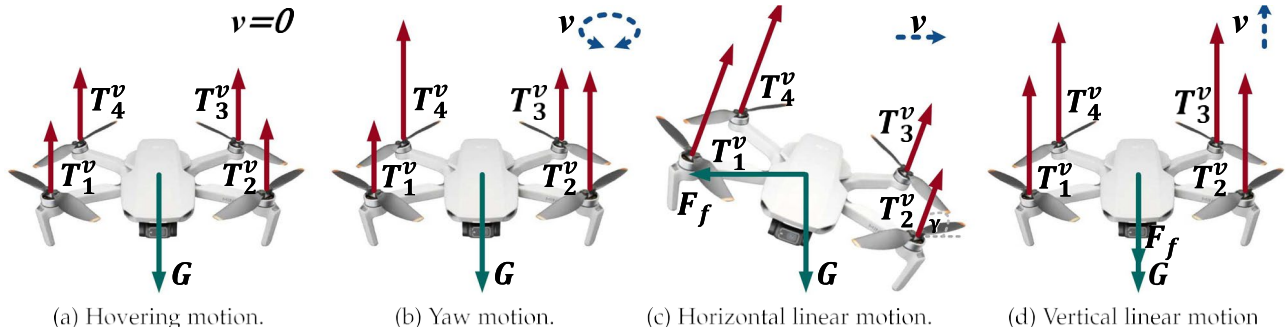

Fig. 5. Force analysis of basic drone motions.

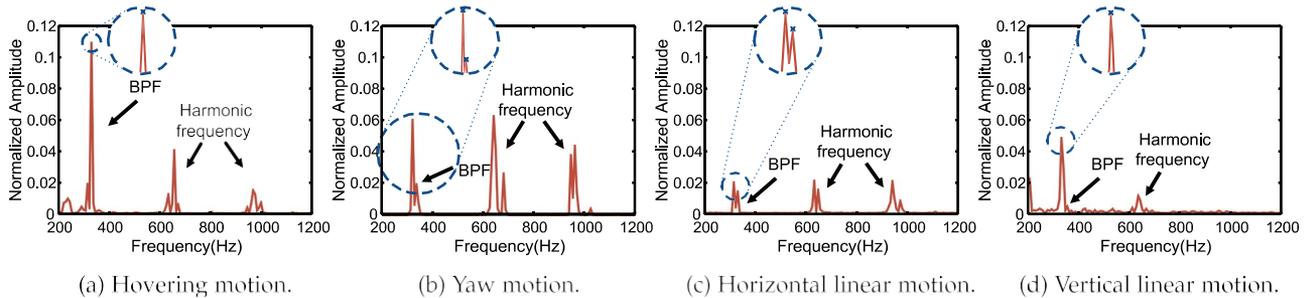

Fig. 6. Acoustic spectrum of basic drone motions.

We design specific band-pass filters for each type of drone, based on their distinctive BPF and harmonic frequencies. For example, the BPF of the DJI Avata, which uses five blades rotating at 300 Hz, is approximately 1500 Hz, while that of the DJI FPV, which uses three blades rotating at 185 Hz, is around 555 Hz, as illustrated in Fig. 7(a) and (b), respectively.

We first process the captured acoustic signal through the band-pass filters of each possible drone structure, to form multiple filtered narrow-band acoustic signals. Then, we calculate the Mel-Frequency Cepstral Coefficients (MFCC) for each filtered signal. MFCC carries information that can effectively represent a drone's sound characteristics in both frequency and time domains [51], so we utilize it to differentiate between drones. Fig. 7(c) and (d) demonstrate the distinct MFCC features of the DJI Avata and FPV, whose energy distributions vary among MFCC vectors, especially where their BPF and harmonic frequencies are located, as shown by the MFCC vectors in red frames.

Finally, we normalize the MFCC vectors of all the filtered signals and borrow the method proposed in DronePrint [51] to train a Long Short-Term Memory (LSTM) neural network for drone identification. If multiple drones are located in the same area, we can identify them according to their corresponding filtered signals.

The profiles of drone structures that cater to a warehouse are pre-archived in a database. Upon identification of a drone, the corresponding profile is fed to dynamics equations for position estimation, which we discuss next.

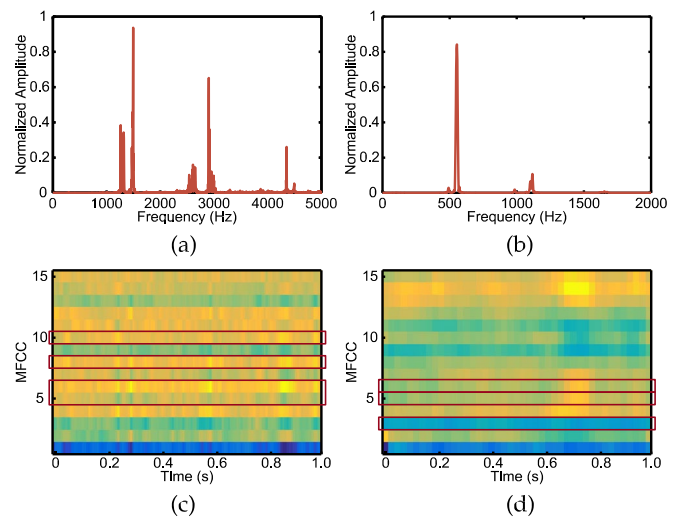

Fig. 7. MFCC of different drones: (a) spectrum of DJI Avata in yaw motion; (b) spectrum of DJI FPV in hovering motion; (c) MFCC of a yawing DJI Avata; and (d) MFCC of a hovering DJI FPV.

## V. DRONE TRAJECTORY TRACKING

We articulate here how to combine information from the drone dynamics with the input from acoustic signals to achieve accurate drone localization and tracking. We further illustrate our system's operation in NLoS settings and how we use a dedicated Kalman filter to tame tracking errors.

## A. Tracking Model

We first derive a dynamic drone model, which we use as a basis for tracking. We consider a quadcopter as an example for intuitive analysis, but the analytical process would be exactly the same for other drone structures.

*Yaw:* In this case, $(T_1^h + T_3^h) - (T_2^h + T_4^h) = 0$, which causes the rotation of the fuselage, as shown in Fig. 5(b), and two BPF peaks. During the rotation process, the moment of inertia $I$ reflects the magnitude of inertia and is regarded as a constant. We can thus obtain the angular acceleration $\beta_t$ at time $t$ by solving the equation:

$$\frac{k_h}{n^2}\left[\sum_{i=1}^{N/2}(f_{2i-1}^{BPF})^2 - \sum_{i=1}^{N/2}(f_{2i}^{BPF})^2\right] = I\beta_t \quad (1)$$

Thus, in a known time interval $\tau$, the rotation angle $\Delta\psi = \int_0^\tau \beta_t\, dt$. However, as mentioned in Section IV-A, ambiguity exists if we only rely on the frequency characteristics. To solve this ambiguity, we regard the drone as a mobile sound source and leverage the microphone array to obtain spatial information. Due to the limited resolution of commercial microphone arrays, the drone is always in the far-field [44], so that we can hardly obtain accurate location information but only a DoA, including azimuth $\alpha$ and elevation $\varphi$. Even in this case, DoA information is sufficient for AIM to function. For instance, DoA information captured by a uniform 4-microphone array in a squared configuration is

$$\begin{cases} \tan\alpha = \dfrac{\tau_{42}^*}{\tau_{43}^*} \\ \sin\varphi = \dfrac{c}{2}\sqrt{\tau_4^{*2} + \tau_3^{*2}} \end{cases} \quad (2)$$

where $c$ is the sound velocity and $\tau^{ij}$ is the time delay between microphones $M_i$ and $M_j$. We calculate the latter with the GCC-PHAT algorithm [52].

*Horizontal motion:* The rotation frequencies of two motors on the same side increase simultaneously to generate a lift force, for example $T_1^v$ and $T_4^v$ in Fig. 5(c), so that the sound contains two groups of BPF peaks, $f_1^{BPF} = f_4^{BPF}$ and $f_2^{BPF} = f_3^{BPF}$. Then the drone tilts with an angle $\gamma$, as shown in Fig. 5(c), so that we can decompose $T_i^v$ into vertical and horizontal directions. The vertical component of $T_i^v$ is balanced with the drone's gravity, so we can solve $\gamma$ with the knowledge of the drone's mass $m$ and the acceleration of gravity $g$, which are known. The horizontal component of $T_i^v$ works against the resistance $F_f = \lambda^h(v_t^h)^2$ to make the drone move horizontally, where $\lambda^h$ can be regarded as a constant related to $\gamma$. We solve the horizontal velocity $v_t^h$ and acceleration $a_t^h$ at time $t$ with the $\gamma$ by the following dynamics equations:

$$\begin{cases} 2\dfrac{k^v}{n}\sum_{i=1}^{N}(f_i^{BPF})^2\sin\gamma = mg \\ 2\dfrac{k^v}{n}\sum_{i=1}^{N}(f_i^{BPF})^2\cos\gamma - \lambda^h(v_t^h)^2 = ma_t^h \end{cases} \quad (3)$$

*Vertical motion:* Consider the case of climbing as an example: $f_i, i = 1, 2, 3, 4$ increase simultaneously to work against the gravity and downward resistance $F_f = \lambda^v(v_t^v)^2$, where $\lambda^v$ can be regarded as a constant, illustrated in Fig. 5(d). Thus, only one BPF peak is captured. Vertical velocity $v_t^v$ and acceleration $a_t^v$ at time $t$ can be determined by solving the equation:

$$\frac{k^v}{n}\sum_{i=1}^{N}(f_i^{BPF})^2 - mg - \lambda^v(v_t^v)^2 = ma_t^v \quad (4)$$

*Finding coordinates:* Consider the situation shown in Fig. 9, where a drone flies from $S_t$ to $S_{t+1}$. A single 4-microphone array with elements $M_1$, $M_2$, $M_3$, $M_4$ is deployed to capture the acoustic signals. The coordinate of the drone at time $t$ are $S_t(h_t\tan\varphi_t\cos\alpha_t, h_t\tan\varphi_t\sin\alpha_t, h_t)$, where the height $h_t$ is now the only unknown quantity. Fortunately, determining $h_t$ is not difficult. For two adjacent coordinates $S_t$ and $S_{t+1}$, in the case of horizontal motion, $h_t = h_{t+1}$, so that

$$|h_{t+1}\tan\varphi_{t+1} - h_t\tan\varphi_t| = v_t^h\tau + \frac{1}{2}a_t^h\tau^2 \quad (5)$$

where $\tau$ is a predefined interval for location updating. In the case of vertical motion, we have

$$|h_{t+1} - h_t| = v_t^v\tau + \frac{1}{2}a_t^v\tau^2 \quad (6)$$

We solve these equations in $h_t$ and determine the complete coordinates of the drone during the flight.

## B. Tracking in NLoS

Indoor scenarios likely include objects that create NLoS settings, for example, in busy warehouses. Here, the DoA information captured by the microphone array may be deviated. For instance, the yellow dashed curves in Fig. 8 depicts the estimated DoA information in NLoS settings. The severe deviation occurs in NLoS no matter whether the drone moves. In this case, traditional triangulation with distributed microphone arrays cannot work, yet alternative indoor localization systems such as UWB- and infrared-based systems may be equally prevented from working altogether in such settings.

In contrast to the state of the art, AIM can recognize if the LoS is blocked and continue to track the drone in NLoS. Despite a few outliers, the dominated diffraction or reflection path with the highest signal energy is stable when the location of the drone is unchanged, while it is irregular when the drone moves. Thus, we employ the Interquartile Range rule (IQR) [53] to eliminate outliers and smooth the estimated DoA information in a sliding window.

When the drone is hovering or yawing, the estimated DoA is smooth, as in Fig. 8(a) and (c), even if the observations slightly deviate from the ground truth. Instead, the smoothed DoA information is erratic when the drone is moving, as in Fig. 8(b) and (d). As described in Table I, we use the stability of DoA information rather than the absolute values to determine the kind of drone motion in LoS. Fig. 8 provides evidence that we can employ the same criteria for the NLoS case.

To detect the NLoS setting in the first place, AIM sets a threshold to evaluate the variance of smoothed azimuth information in a time window. If the variance is beyond the threshold, we consider the LoS to be blocked, because even if smoothed, the DoA in NLoS is still unstable, which is especially evident in azimuth estimation, as shown by the green curve in Fig. 8(b).

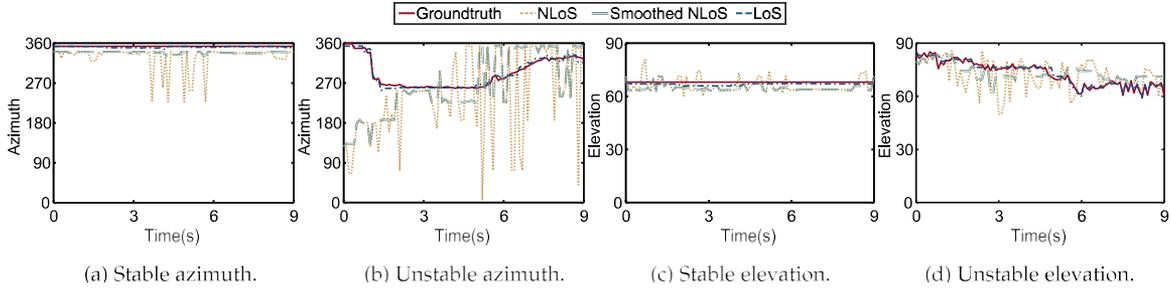

(a) Stable azimuth.    (b) Unstable azimuth.    (c) Stable elevation.    (d) Unstable elevation.

Fig. 8. DoA estimation results in LoS and NLoS.

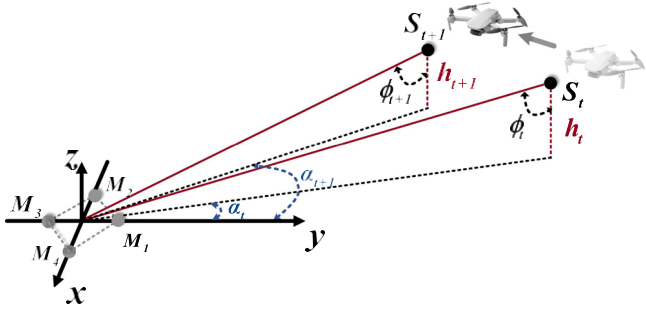

Fig. 9. Schematic diagram of AIM in action.

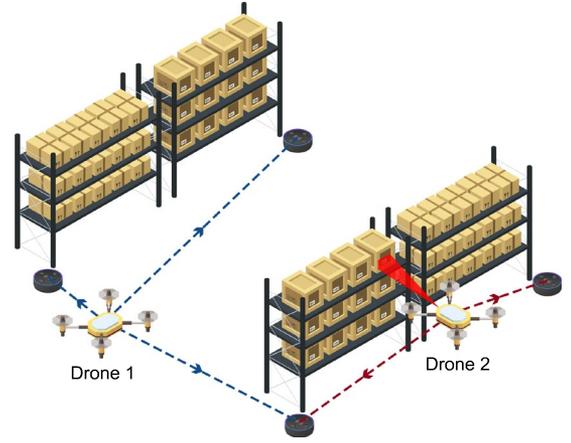

Fig. 10. Example scenarios of tracking with multiple arrays.

### C. Error Calibration

We employ a dedicated Kalman filter to tame the inaccuracies in the estimation of orientation after yawing and in absolute localization following horizontal or vertical motion.

The drone location is described by a state vector $A_t = [x_t, y_t, z_t]^T$, with $A_0$ being initialized with the first few points at the beginning of the flight. Then processing unfolds as follows:

1) We predict the subsequent state vector $\hat{A}_t^-$, that is, the a priori state estimate, according to the state transition matrix;
2) We estimate the drone's current motion following the rules in Table I as well as the current coordinate according to the dynamic equations and identified motion;
3) Based on the variance of the smoothed azimuth, we identify whether the LoS exists. If not, the estimated DoA information is discarded;
4) With yaw motion, possible trajectories caused by the ambiguous orientations are tracked until the LoS is regained. If the LoS exists now, the current coordinates can be updated with DoA, eliminating the ambiguity;
5) No matter whether in LoS or NLoS, the measured coordinates are fused with $\hat{A}_t^-$ to output the optimal estimate $A_t$, that is, the a posteriori state estimate.

## VI. EXTENDING OPERATING RANGE

Despite the ability of AIM to operate with a single microphone array, in realistic indoor settings such as a warehouse, the coverage may be insufficient. As a result, we extend our tracking scheme using distributed microphone arrays to accommodate indoor environments with variable ranges and configurations.

### A. Basic Model

An example of a warehouse employing distributed microphone arrays is depicted in Fig. 10. In this scenario, the arrays are positioned at regular intervals among the shelves to facilitate tracking of the drone through relaying. As the drone traverses these zones, we use neighboring microphone arrays to calculate its location and subsequently refine the results reported by acoustic inertial measurement, thereby enhancing localization accuracy.

Our approach involves computing the time difference of arrival (TDoA) between each pair of microphone arrays. We uniformly orient all arrays in the same direction and number their elements according to consistent rules. If we designate the $m$-microphone arrays $Arr_p$ and $Arr_q$ to have elements $M_1^p\ldots M_m^p$ and $M_1^q\ldots M_m^q$, respectively, the TDoA $T_{pq}$ between the two arrays is determined as:

$$T_{pq} = \frac{\sum_{i=1}^{m} \tau^*(M_i^p, M_i^q)}{m} \quad (7)$$

where $\tau^*(M_i^p, M_i^q)$ is the time delay between the corresponding elements of two arrays.

It follows that the locations of the drone that satisfy this TDoA form a hyperboloid, as depicted in Fig. 11. Here, we denote the drone's location at time $t$ as $S_t(x_t, y_t, z_t)$ and the positions of the two arrays as $Arr_p(x_p, y_p, 0)$ and $Arr_q(x_q, y_q, 0)$. The shape of the hyperboloid is derived from the calculated TDoA as follows:

$$F(Arr_p, Arr_q) = \frac{x^2}{a^2} - \frac{y^2}{b^2} - \frac{z^2}{c^2} - 1 \quad (8)$$

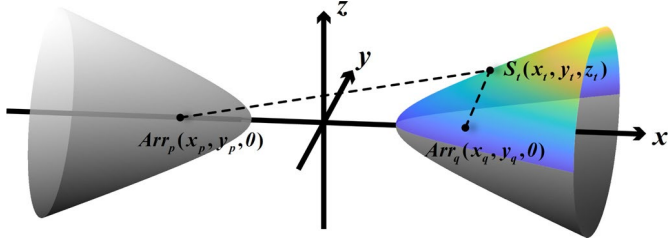

Fig. 11. TDoA between two microphone arrays.

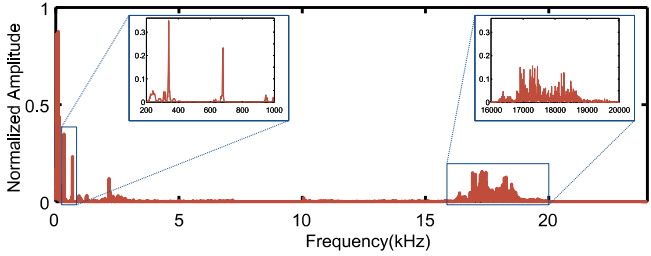

Fig. 12. Spectrum when the beacon and drone sound exist simultaneously.

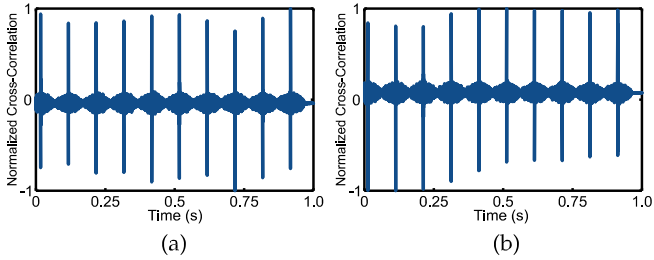

Fig. 13. Detecting the presence of the beacon: (a) when the drone does not take off; (b) when the drone is hovering.

where $a = \frac{1}{2} \cdot abs(||S_t\ Arr_q|| - ||S_t\ Arr_p||) = \frac{1}{2} c \cdot T_{pq}$ and $b = c = \sqrt{\frac{1}{2} \cdot ||Arr_p\ Arr_q||^2 - a^2}$.

With at least three microphone arrays, say $Arr_p$, $Arr_q$, and $Arr_s$, we can estimate the drone's location at time $t$ by solving the following set of equations:

$$S_t(x_t, y_t, z_t) = \begin{cases} F(Arr_p, Arr_q) = 0; \\ F(Arr_p, Arr_s) = 0; \\ F(Arr_q, Arr_s) = 0; \end{cases} \quad (9)$$

To synchronize the microphone arrays involved in location estimation, we employ commercial speakers to intermittently emit an acoustic beacon, which consists of a pre-defined pseudo-random noise. During drone tracking, the microphone arrays detect this beacon to align with one another [54]. The beacon frequency ranges from 16 kHz to 20 kHz, as depicted in Fig. 12, and is distinct from the signals used for localization, making it separable via band-pass filters. As shown in Fig. 13(a) and (b), the beacon is accurately detected also when the drone is present.

There may be cases where a drone can establish line of sight with only two microphone arrays, as for drone 2 in Fig. 10. If so, (9) becomes negative definite or non-full rank, rendering it unsolvable and it becomes impossible to obtain the 3D coordinates of the drone. To address this issue, we no longer treat each microphone array as a whole, as in (8), and instead choose multiple individual microphone elements for localization.

Say we can only rely on two microphone arrays.[1] In this case, we choose two elements from each array, respectively, and consider each of them as a new two-microphone array. The distance between two elements in the same array must be the largest, and furthermore, the selected four elements must not be collinear. We can then employ the model in Section VI-A to calculate the drone's location using the four selected microphone elements.

### B. Selecting Microphone Arrays for Localization

Although there may be several microphone arrays located near the drone that can receive the acoustic signal with a high amplitude, some of them may be in NLoS or surrounded by multiple reflectors. If these arrays are chosen to perform TDoA and location calculation, the resulting localization information may be inaccurate. To mitigate this issue, we execute a dedicated array selection algorithm, which is depicted in Algorithm 1.

We perform a preliminary screening using the method outlined in Section V-B to filter out microphone arrays that report unstable or inaccurate results. If the number of the remaining arrays is enough to determine the 3D coordinates of the drone, that is, there are at least 3 arrays that show reliable DoA estimation, we proceed to the following fine-grained selection process. Otherwise, if all arrays are in a line or only two arrays can be used, we calculate the location as described above.

Next, we apply an additional filtering process to further refine the selected microphone arrays. Let $d_{pq}$ denote the distance between two microphone arrays $Arr_p$ and $Arr_q$. We select the first three microphone arrays as the initial set, where the product of the distance between them is the largest. This is because TDoA estimates tend to be more accurate when microphone arrays are more dispersed.

With the selected three microphone arrays, we obtain an initial estimation of the drone's coordinates. This estimation may not be stable enough as it is based on only three microphone arrays. If there is any other candidate microphone array providing preferable DoA estimation, we add this to the processing to improve the accuracy, choosing the one with the most stable DoA estimations. To further enhance the accuracy, results reported by the distributed microphone arrays are fused with those obtained from acoustic inertial measurement, as explained next.

### C. Fusing Data

After obtaining the drone location from (9), we fuse this result with that of the acoustic inertial measurement. We use the complementary filter for this, because of two reasons. First,

---

[1] If there are multiple arrays arranged on a single line, we select the two nearest to the drone.

**Algorithm 1:** Array Selection Algorithm.

```
1  for t = 1, 2, 3, … do
2      Determine the set of microphone arrays C_Arr;
3      if COUNT(C_Arr) > 3 then
4          for Arr_p ∈ C_Arr do
5              if Variance Var_α̂_p < Threshold then
6                  Add Arr_p to the candidate set
                   C_selected;
7              end
8          end
9          Load the distance d_pq between every
           microphone array in C_selected;
10         Initial set C_initial = arg max d_pq * d_ps * d_qs;
                                  (p,q,s)
11         for Arr_p ∈ C_selected − C_initial do
12             Find the Arr'_p with the minimum Var_α̂_p;
13         end
14         Add Arr'_p to C_initial;
15     end
16     Output C_initial;
17 end
```

location estimations output by distributed microphone arrays can exhibit jitter, which results in high-frequency noise, while estimations of acoustic inertial measurement are smooth in a short period of time, which can therefore effectively compensate for this problem. On the other hand, both distributed microphone arrays and acoustic inertial measurements produce fairly accurate results so we can employ the lightweight complementary filter to avoid nesting of two Kalman filters, greatly reducing processing times.

Let $s(\Delta t)$ denote the true trajectory of the drone over a time period $\Delta t$, so we have

$$z_M(\Delta t) = s(\Delta t) + n_1(\Delta t)$$
$$z_A(\Delta t) = s(\Delta t) + n_2(\Delta t) \quad (10)$$

where $z_M(\Delta t)$ and $n_1(\Delta t)$ are the estimation results and noise of the distributed microphone arrays, and $z_A(\Delta t)$ and $n_s(\Delta t)$ are those of the acoustic inertial measurement. Then we perform a data fusion process based on

$$\hat{S}(f) = Z_M(f)G(f) + Z_A(f)[1 - G(f)] \quad (11)$$

where $\hat{S}(f)$ is the Fourier transform of the fused result $\hat{s}(\Delta t)$, $Z_M(f)$ and $Z_A(f)$ are the Fourier transform of $z_M(\Delta t)$ and $z_A(\Delta t)$, and $G(f)$ and $1 - G(f)$ is the low-pass filter and the complementary high-pass filter.

Finally, we can obtain the fused result $\hat{s}(\Delta t)$ by performing inverse Fourier transform for $\hat{S}(f)$.

## VII. EVALUATION

We report evaluation results of AIM using off-the-shelf microphone arrays and a commercial drone. We describe first the implementation and evaluation settings in Section VII-A. Next, our investigation of AIM performance is two-pronged: Section VII-B compares our system with the state-of-the-art indoor drone tracking systems and reports on their performance under different scenarios; in Section VII-C, we dissect the impact on tracking accuracy of environment noise, flight range and velocity, as well as of the deployment configurations of distributed microphone arrays and of the beacon volume. We discuss the real-world performance of AIM in Section VII-D.

Our results indicate that:
1) The mean localization error of AIM in NLoS settings, arguably most realistic for indoor drone applications, is 46% lower than a UWB-based baseline;
2) Unlike an infrared-based baseline, AIM constantly provides location updates, even in NLoS settings;
3) AIM is robust to moderate noise sources in the environment, such as someone speaking;
4) Flight range and velocity of the drone influence AIM's performance differently, yet the absolute accuracy never degrades drastically.
5) With distributed microphone arrays, AIM can be extended to support indoor spaces with arbitrary ranges and layouts without loss of accuracy.

### A. Implementation and Settings

AIM works with any layout of bidimensional microphone array to track drones of various structures. Without loss of generality, here we consider a quadcopter and two types of microphone arrays.

*Drones and microphone arrays:* We use a DJI Mini 2 quadcopter [55], shown in Fig. 14(a). The DJI Mini 2 weighs 249g; as such, flying the DJI Mini 2 in most countries does not require a professional drone piloting license, which makes it ideal for indoor use. Each propeller is equipped with two blades. When the drone is hovering, the sound pressure level measured at a 1m distance is empirically determined to be around 77 dB and motors run at 164 Hz, so the BPF is around 328 Hz. By default, the DJI Mini utilizes the built-in GPS for horizontal localization and an infrared time of flight (ToF) sensor to obtain vertical altitude. However, in the indoor experimental environment we use, shown in Fig. 14(b), GPS cannot work and only the ToF sensor provides useful altitude information.

We use two types of commercial off-the-shelf microphone arrays for our AIM prototype: a Seeed Studio ReSpeaker 6-mic circular array [56] and Seeed Studio ReSpeaker 4-mic array [57], shown on the upper left of Fig. 14. The inter-distance between two single microphones is 5 cm and 6.5 cm, respectively. Each microphone array is set on a Raspberry Pi 4 Model B, using a 48 kHz sampling rate. Unless stated otherwise, the results we discuss next are obtained with the 6-mic circular microphone array.

*Baselines:* To obtain ground-truth information, we take the readings of the built-in ToF sensor on the DJI Mini 2 as vertical altitude. As for the horizontal coordinates, we employ a method often used in indoor drone testbeds [58]: we lay down distance markers on the ground at intervals of 10 cm, as shown in Fig. 14(b) and (c). Using the downward-facing camera of the drone, we examine its view of the ground-level markers during the flight. Fig. 14(c) shows an example image captured by the drone during the experiments. Once the tick of the marker matches the centerline of the image, this reading of the corresponding maker is regarded as the real-time horizontal coordinates.

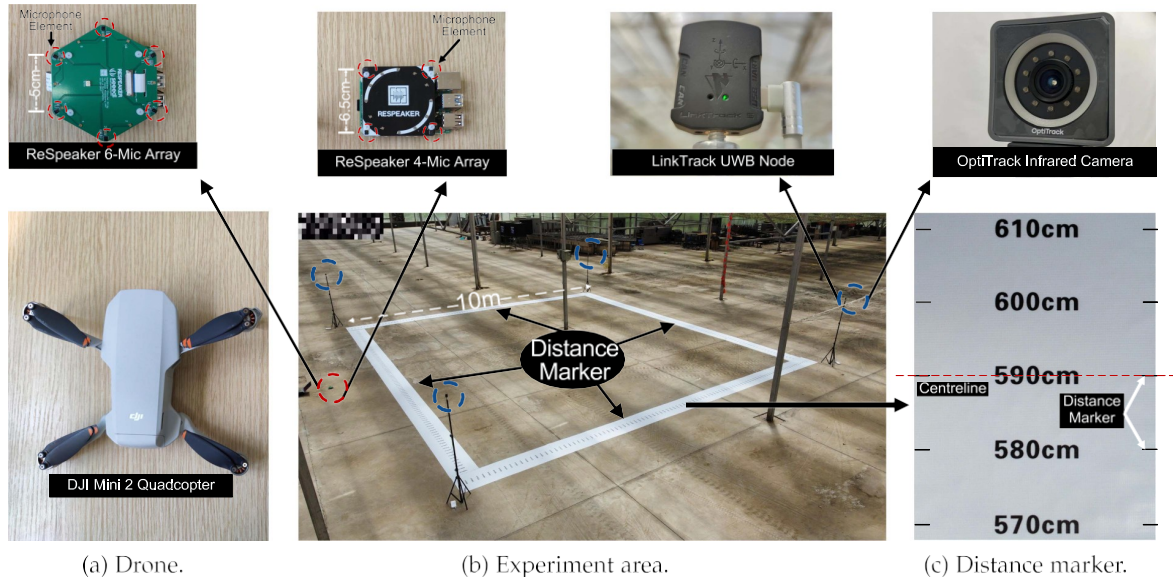

Fig. 14. Experiment settings.

We compare AIM with LinkTrack [59], an UWB-based indoor localization system, and OptiTrack [38], an infrared-based motion tracking system, both of which are shown on the upper right of Fig. 14. LinkTrack localizes the target via triangulation. We fix a UWB tag on the drone and four UWB anchors on four tripods, then record the tracking results on a base station. OptiTrack localizes the target by converting the drone positions in bidimensional photos captured at high frequency by multiple infrared cameras to three-dimensional coordinates. We fix reflective markers on the drone and four infrared cameras on four tripods, and also record the tracking results on a base station. Whenever the drone carries a UWB tag or reflective markers, we accordingly update its tracking model and dynamic parameters.

Note that the OptiTrack system is vastly considered as state of the art in indoor testbeds. Because of its cost, difficulty in installation, and inability to work in NLoS settings, however, it is rarely employed for real applications [58].

*Scenarios and drone mobility:* We select three scenarios. In *Line-of-Sight* (LoS), nothing is deployed in the middle of the experiment area shown in Fig. 14(b) and every device involved in localization can establish LoS with each other and with the drone. Note how this scenario, while common in indoor drone testbeds that are in fact designed to isolate drones from their surroundings, is quite unlikely in real applications. In *Partial Line-of-Sight* (PLoS), several steel shelves stacked with various objects such as books and bricks are deployed in the middle of the experiment area. As shown in Fig. 15(a), depending on the relative position of the drone with respect to the rest of the experiment area, the LoS is blocked at times. In *None-Line-of-Sight* (NLoS), the shelves are deployed in front of every tripod hosting infrastructure node for localization. Every LoS path is thus blocked, as depicted in Fig. 15(b). No matter where the drone flies in the experiment field, it can not establish LoS connection to any device on any of the tripods.

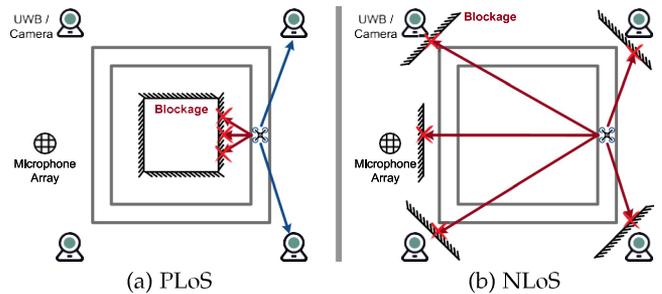

Fig. 15. Experiment scenarios.

We tested varied combinations of drone motions. For *horizontal motions*, we control the drone to fly along the distance maker, shown in Fig. 14(c), and keep vertical coordinates unchanged. For *vertical motions*, once the drone is hovering, we control the drone to climb or descent to a certain height, while keeping horizontal coordinates unchanged.

### B. General Performance

We fly a 10m×10m squared trajectory comparing AIM with LinkTrack and OptiTrack in LoS, PLoS and NLoS scenarios. Fig. 16 reports the performance of the three systems.

Fig. 16(a) indicates that in LoS scenarios, the mean error of AIM is 1.43m while those of LinkTrack and OptiTrack are 0.37m and 0.03m, respectively.[2] AIM is, therefore, the least accurate

---

[2]Note that for OptiTrack, we note a difference between the error measured in our experiments and what is advertised by the manufacturer, which is below 1mm. The reason for this is that OptiTrack sometimes temporarily recognizes LEDs on the drones as the markers, affecting the measurements. We cannot turn off or cover these LEDs, as the drone would refuse to take off, raising exceptions in the control software.

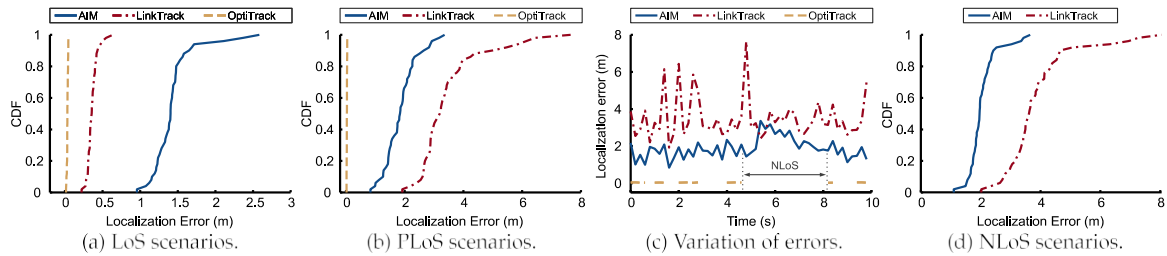

Fig. 16. Performance comparison.

system in LoS scenarios, which are, however, arguably rare in real applications.

Fig. 16(b) illustrates the performance in PLoS scenarios. Here AIM outperforms LinkTrack with a mean error of 1.89 m, which is 46% less than LinkTrack. The increase of error is caused by the lack of DoA calibration for AIM and by signal attenuation for LinkTrack. In that case, AIM can only calibrate the location with the opportunistic clean DoA.

Fig. 16(c) offers a closer view on this specific experiment by showing an accuracy comparison during a 10 s flight, including about 2 s of NLoS. LinkTrack is heavily influenced by the obstacles, which absorb UWB signals. When the LoS is obstructed, OptiTrack simply does not work and produces no output. Thus, although its mean error does not increase in PLoS scenarios, OptiTrack is plainly inapplicable as completely losing the drone position even for a short among of time would be unacceptable for safe and dependable operation. Instead, the localization error of AIM suddenly increases at the beginning of the NLoS sting, but gradually decreases later, without ever losing the target.

In NLoS scenarios, shown in Fig. 16(d), we only compare AIM with LinkTrack because OptiTrack produces no output for the entire duration of the experiments, because of the aforementioned reasons. The mean error of AIM increases to 2.08 m but it is still lower than that of LinkTrack, which is almost twice as much at around 4 m.

Note how the progression through different scenarios in our discussion, from LoS in Fig. 16(a) to NLoS in Fig. 16(d), reflects increased realism in indoor drone applications. NLoS settings are indeed expected to abound when drones fly in complex physical environments. These settings are precisely where AIM reaps the greatest benefits compared to the baselines: its performance degradation, indeed, is much less pronounced compared to LinkTrack, whereas it can supply continuous location updates, unlike OptiTrack.

### C. Factors Influencing Accuracy

We analyze the impact of three different factors on localization accuracy, that is, noise in the environment, the flight range and velocity, and the number of microphones.

*Environment noise:* We examine the performance of AIM in noisy conditions. We place a noise source 2m away from the microphone array. To study different degrees of interference, we set the volume of the noise source to 50 dB, 55 dB, 60 dB and 65 dB. We broadcast Gaussian white noise with 100 Hz bandwidth in three different center frequencies, that is, at 300 Hz, 600Hz and 900 Hz, to simulate interference on the BPF and its harmonic frequency.

The results in Fig. 17 indicate that, as expected, the localization accuracy degrades as the frequency of the noise or the SPL of the noise increases. This is because AIM weights the BPF and its harmonics according to their amplitude and sums them up to obtain the final frequency, which is the input of dynamic equations. In general, BPF and lower harmonics exhibit higher energy and thus are given higher weights. However, if the noise is at high frequency, peaks in this frequency band gain much higher weights. Therefore, the results are polluted.

Importantly, results show that AIM still maintains relatively stable performance under noisy conditions, which is sufficient to deal with common noise environments such as someone speaking, which is around 53.7 dB at 1 m distance. We also demonstrate that AIM can cope with narrowband noise, whose frequency band does not violate all the BPF and harmonic frequencies simultaneously. Even faced with broadband noise (e.g., music), AIM still provides accurate localization results as long as the noise intensity is lower than that of the drone signals. If not, multiple options exist to resist noise in practice. We may, for example, introduce a band-pass filter to filter out the noise band and continue tracking using the uncontaminated frequency band. AIM is also flexible in the deployment of the microphone array, as no specific requirements must be fulfilled to during installation. We may simply alter its position to lessen the impact of nearby noise sources.

*Flight range and velocity:* First, we investigate the performance of AIM depending on the distance between the drone and the microphone array. We specifically test three flight paths, composed of 5 m $\times$ 5 m, 10 m $\times$ 10 m, and 15 m $\times$ 15 m square trajectories. The drone is controlled to fly at a velocity of 1.5 m/s in both horizontal and vertical motions. Fig. 18 shows the results.

When the drone flies along the 5m$\times$5m square, the mean errors are 0.95 m in LoS and 1.52 m in PLoS. When the drone flies along the 10 m $\times$ 10 m square, the mean errors are 1.43 m in LoS and 1.89m in PLoS. If the drone flies over a larger area, the signal attenuation worsens so the error increases. Correspondingly, the results show that the mean errors in both LoS and PLoS are over 2 m as the drone flies along a 15 m $\times$ 15 m field.

Based on these results, we define 10 m as the *operational range* for the pair DJI Mini 2/ReSpeaker 6-mic. The operational range is an empirical value, which sets a limit on the acceptable tracking error. Note that this value may be different between different drones and microphone arrays, as it is mainly determined by the SPL of the sound produced by the drone's propellers and the sensitivity of the microphone array. The higher the drone's SPL and the array's sensitivity, the lower the tracking error in a given field and the larger the operational range.

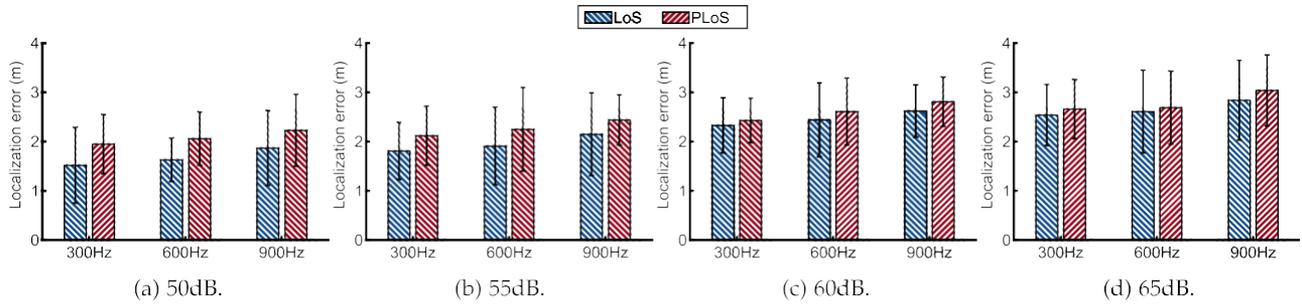

Fig. 17. Impact of environment noise on accuracy.

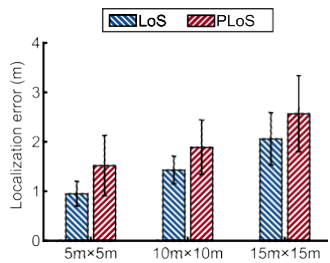

Fig. 18. Flight range.

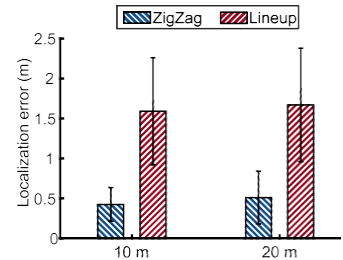

Fig. 20. Deployment of mics.

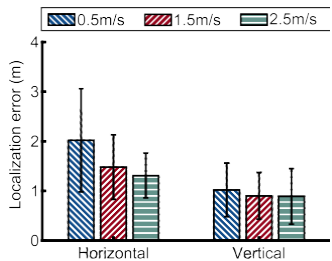

Fig. 19. Flight velocity.

We also conduct experiments to evaluate if the drone's velocity has an impact on accuracy. These experiments are conducted in the LoS scenario, and both horizontal and vertical motion are evaluated, respectively. In the horizontal motion, we control the drone to fly along a 10m×10m square. The results are shown in Fig. 19. For horizontal motion, the drone's velocity influences the accuracy in that the mean error decreases as the velocity increases, while for vertical motion, the change of velocity does not significantly impact accuracy. The reason is two-fold. On the one hand, two frequency peaks must be captured for horizontal motion. Higher velocity results in larger intervals between the two frequency peaks, hence they are easier to separate out. In contrast, only one peak must be captured during vertical motion. On the other hand, every two propellers contribute to the energy of one frequency peak with horizontal motion, while all propellers generate the signal at the same frequency with vertical motion. The energy of the frequency peak in vertical motion is higher than that in horizontal motion and, therefore, results in more stable performance.

*Deployment of microphone arrays and beacon volume:* We evaluate the localization accuracy in continuous drone tracking by varying the deployment of microphone arrays and the volume of the beacon.

Firstly, we deploy several microphone arrays in two different configurations: ZigZag and straight lines. Then, we compare the localization accuracy of these two deployments in the 10m and 20m range. In the straight line setting, we place several arrays in a line, and to simulate the corner of the warehouse, we also place one array at the end of the line that is not colinear with the others. This arrangement provides the opportunity to perform error calibration with at least three microphone arrays. In the ZigZag setting, the arrays are placed in two lines as a form of ZigZag and the distance between the two lines is 10 m. The drone is controlled to fly along the center line of two lines, so the horizontal distance between the drone and each microphone is around 5 m. The drone velocity is 1.5 m/s and all microphone arrays can establish a LoS with the drone.

The results in Fig. 20 show that the ZigZag configuration provides much better accuracy, with errors less than 0.5 m, in both the 10 m and 20 m range. As the horizontal distance between the drone and each microphone array during flight is around 5 m, and the flight height is 2 m, the relative error in this setting is less than 9.28% ($0.5/\sqrt{5^2 + 2^2}$). In contrast, the errors in the straight line setting are around 1.5 m, even with the opportunity for calibration. Thus, we recommend deploying distributed microphone arrays as in the ZigZag configuration for better performance, if conditions permit.

We also investigate the impact of varying the volume of the time synchronization beacon. The experiments are conducted with microphone arrays deployed in the ZigZag configuration, while the drone flies in the 10m range with the velocity of 1.5 m/s. As Fig. 21 shows, increasing the volume of the beacon leads to a reduction in localization error. Specifically, the error decreases from 1.43 m at 40 dB to 0.45m and 0.44 m, at 60 dB and 70 dB, respectively. However, noise can also exist in the band of the beacon, even with high frequency, and therefore, higher volumes may not always result in better performance. Moreover, some industrial settings may have strict regulations on sound volume, including those in the frequency range that is not audible to humans. To address these limitations, we may extend the length of the beacon, instead of increasing the volume, which can compensate for the reduction in volume without affecting the performance.

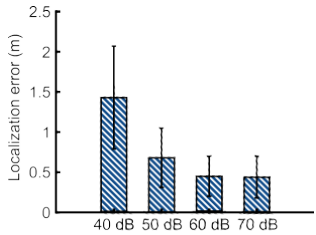

Fig. 21. Volume of beacon.

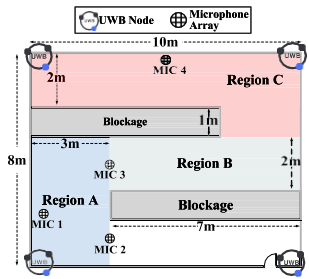

Fig. 22. Warehouse layout.

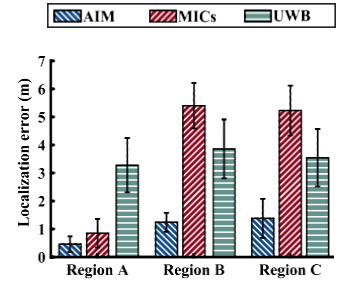

Fig. 23. Accuracy.

### D. Performance in Realistic Settings

We offer further evidence on the real-world applicability of AIM. The instrument we use to this end is a real deployment in a warehouse, whose layout is shown in Fig. 22. At three different regions in the warehouse we compare the localization accuracy of AIM with that of LinkTrack and triangulation using distributed microphone arrays, which is usually used in many acoustics-based localization methods [60], [61], [62].

Fig. 23 reports the results. In Region A, triangulation achieves a fair accuracy with a mean error of 0.85 m. In comparison, AIM reports shows more accurate results with a mean error of 0.46 m. The reason is that AIM can fuse the results from distributed microphone arrays to output more precise and stable results. When the drone enters Region B and Region C, triangulation becomes inapplicable, as it returns an error above 5 m, but AIM's performance is not affected. This is because our system only requires one LoS to disambiguate or not even that, whenever the drone does not perform yaw motion in NLoS. In contrast, for triangulation to work, LoS from all microphone arrays is mandatory.

As for LinkTrack, we set the four UWB anchors at the corners of the area to cover the whole warehouse, as shown in Fig. 22. In such a deployment configuration, LinkTrack performs poorly in all three regions because of the signal loss caused by the obstacles in the warehouse.

## VIII. DISCUSSION

We complete the discussion of AIM by articulating practical issues of applicability and general use. Two aspects are worth considering here.

*Sensor fusion for indoor tracking:* Different techniques have their unique advantages and disadvantages. Multiple techniques could be combined to improve performance.

Most existing commercial drones are already equipped with multiple sensors, including ToF, IMU and cameras, for accurate indoor localization. In the context of passive drone tracking, sensor fusion is also feasible. For example, one may deploy UWB nodes or cameras at the corner to calibrate the drone's location, while exploiting microphone arrays in other places to reduce cost. Besides, in PLoS indoor scenarios like Fig. 15(a), the drone can establish LoS paths with at least two sensors in most cases. Therefore, a real-time sensor fusion algorithm can be applied to achieve accurate localization results. However, strict time synchronization between different sensors and quick identification of LoS paths are required.

*Multi-drone tracking:* When multiple drones enter the same area, AIM can still track them separately if their BPF are different. Otherwise, frequency aliasing happens. We may handle this problem by borrowing ideas from existing works to discriminate different sound sources along different propagation paths [44] or to modulate the unique acoustic signature in the drone motor sound [18].

## IX. CONCLUSION

We presented AIM, a one-of-a-kind passive indoor drone tracking technique that works with a single microphone array, but may also be extended to support spaces with any range and layout by deploying distributed microphone arrays. AIM innovates the acoustic tracking technique in that it fully exploits the dual acoustic channel from the drone to the microphone array, based on an in-depth understanding of the drone's dynamics and the characteristics of its acoustic signal. Through extensive experiments, we demonstrate that AIM offers strikingly better performance than state-of-the-art solutions, especially in NLoS settings, and enjoys stable performance across complex indoor environments.

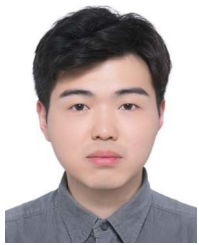

**Weiguo Wang** (Student Member, IEEE) received the BE degree from the University of Electronic Science and Technology of China, and the PhD degree from Tsinghua University. His research interests include acoustic sensing and mobile computing.

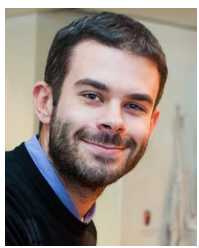

**Luca Mottola** (Member, IEEE) is currently a full professor with the Politecnico di Milano, Italy, and a senior researcher with RISE Sweden. He is the past general chair for ACM/IEEE CPS-IoT Week 2022 and past PC chair for ACM MOBISYS, ACM SENSYS, ACM/IEEE IPSN, and ACM EWSN. He was the recipient of the ACM SENSYS Test of Time Award in 2022, two ACM SigMobile Research Highlights, and is a Google Faculty Award winner. He holds or held visiting positions with Uppsala University, NXP Technologies, TU Graz, and USI Lugano.

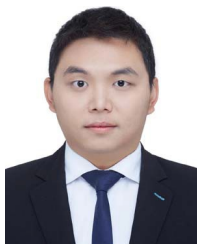

**Jia Zhang** (Graduate Student Member, IEEE) received the BE degree from Tsinghua University, in 2019. He is currently working toward the PhD degree with the School of Software and BNRist, Tsinghua University. His research interests include Internet of Things and wireless sensing.

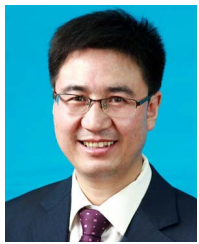

**Ruijin Wang** (Member, IEEE) is currently an Associate professor of University of Electronic Science and Technology of China. He is the secretary general of ACM Chengdu Chapter, senior member of CCF and member of Asia Pacific Association for Artificial Intelligence (AAIA). He is a visiting scholar with Northwestern University. His research interests include cloud-edge intelligent computing, artificial intelligence, security, and blockchain.

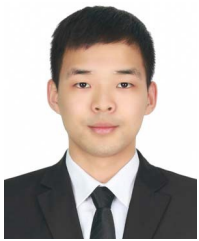

**Yimiao Sun** (Student Member, IEEE) received the BE degree from the University of Electronic Science and Technology of China. He is currently working toward the PhD degree with Tsinghua University. His research interests include mobile computing and wireless sensing.

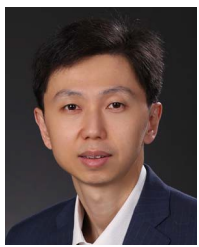

**Yuan He** (Senior Member, IEEE) received the BE degree from the University of Science and Technology of China, the ME degree from the Institute of Software, Chinese Academy of Sciences, and the PhD degree from Hong Kong University of Science and Technology. He is currently an associate professor with the School of Software and BNRist, Tsinghua University. His research interests include wireless networks, Internet of Things, pervasive and mobile computing. He is a member ACM.